\documentclass[11pt]{article}
\usepackage{coling2020}
\usepackage{coling-nert}
\usepackage{times}
\usepackage{url}
\usepackage{latexsym}

\colingfinalcopy 

\title{PASTRIE: A Corpus of Prepositions Annotated with Supersense Tags in Reddit International English}

\author{Michael Kranzlein \quad Emma Manning \quad Siyao Peng \\ {\bf Shira Wein} \quad {\bf Aryaman Arora} \quad {\bf Bradford Salen} \quad {\bf Nathan Schneider}\\
Georgetown University \\
\{\emldisplay{mmk119@georgetown.edu}{mmk119},
\emldisplay{esm76@georgetown.edu}{esm76}, 
\emldisplay{sp1184@georgetown.edu}{sp1184},
\emldisplay{sw1158@georgetown.edu}{sw1158},
\emldisplay{aa2190@georgetown.edu}{aa2190},
\emldisplay{bls85@georgetown.edu}{bls85},
\emldisplay{nathan.schneider@georgetown.edu}{nathan.schneider}\}\texttt{@georgetown.edu}
}

\date{}

\begin{document}
\maketitle

\begin{abstract}

We present the Prepositions Annotated with Supersense Tags in Reddit International English (``PASTRIE'') corpus, a new dataset containing manually annotated preposition supersenses of English data from presumed speakers of four L1s: English, French, German, and Spanish. 
The annotations are comprehensive, covering all preposition types and tokens in the sample.
Along with the corpus, we provide analysis of distributional patterns across the included L1s and a discussion of the influence of L1s on L2 preposition choice.

\end{abstract}

\section{Introduction}\label{intro}
It is well-established that one's native language (``L1'') leaves traces in second language (``L2'') word choice and grammar, including subtle aspects of the use of function words such as prepositions---even for highly proficient L2 speakers \cite{lowie2004,kujalowicz2005,mueller2011,Nacey_Graedler_2015}. However, past corpus studies of L2 writing have had no way to control for the \emph{meaning} of these grammatical items in context on a large scale. 
\blfootnote{
    \hspace{-0.65cm}  
    This work is licensed under a Creative Commons
    Attribution 4.0 International License.
    Licence details:
    \url{http://creativecommons.org/licenses/by/4.0/}.}
In this work, we describe a new corpus, PASTRIE,\footnote{PASTRIE is available at \url{https://github.com/nert-nlp/pastrie}} consisting of English Reddit posts and comments (collectively ``documents'') that have been manually annotated with preposition supersenses.%
\footnote{Automatic lemmas and part-of-speech tags are also included.}
Following \citet{schneider2018}, the annotations also cover possessives, prepositional multiword expressions (``MWEs''), and infinitives.\footnote{We may sometimes use the word ``preposition'' loosely to cover all of these categories. When specific analyses are being made, more precise terminology is used.}

Examples of annotated sentences appear in \cref{ex:intro1} and \cref{ex:intro2}.

\ex.\label{ex:intro1} I was just \p{on}/\psst{Locus} it \p{to}/\psst{Purpose} find the Copenhagen deal and couldn't find it \p{at\_first}/\psst{Time} .

\ex.\label{ex:intro2} Right \p{at}/\psst{Time} the moment when that geyser \p{of}/\psst{Stuff} light erupts \p{from}/\psst{Source} the edge \p{of}/\psst{Whole} the screen , we hear a massive rumble come \p{from}/\psst{Source} the door , which was \p{in}/\psst{Locus} that direction .

The annotations are comprehensive, covering all types and tokens of prepositional expressions, totaling 2400 tokens out of the  22.5k token corpus.
The documents are drawn from the larger Reddit-L2 corpus \cite{rabinovich2018}, which consists of English Reddit data of speakers of many different L1s. Our corpus includes English produced by presumed native speakers\footnote{Information on how L1s were identified is included in \cref{sec:description}. Following \citet{rabinovich2018}, we simply say ``native speakers'' or ``L1'' with the understanding that this is an imperfect assumption.} of English, French, German, and Spanish.

Based on annotators' impressions, the English in the corpus produced by the nonnative speakers is highly fluent and unlike what might be found in learner corpora. This is understandable given that these users are taking it upon themselves to post in an online forum, something early learners are less likely to do. This corpus is not only a new resource for exploring preposition supersenses, but it also addresses an understudied niche of broad-coverage semantics for highly proficient non-native data.
Using a large, unannotated sample of the Reddit-L2 corpus as well as our semantically-annotated subcorpus, we conduct a preliminary investigation of preposition use among English speakers of different L1 backgrounds,
extending \citeposs{rabinovich2018} analysis of L2 lexical choice.
We will release the corpus to facilitate further study of such phenomena.

\section{Related Work}
\subsection{Preposition Supersenses}
Supersenses are categories used to place both content and function words into unlexicalized semantic classes~\cite{schneider2015}, and have been applied to nouns, verbs, adjectives, and adpositions\footnote{Adpositions include prepositions, postpositions, and circumpositions, but since we are concerned with English data, we often only mention ``prepositions.''} \cite{miller1990,fellbaum1990,tsvetkov2015}. Here, we focus on the latter. Though adpositions (which almost always occur as prepositions in English) are considered function words and often treated as less important in natural language processing contexts, \shortcite{schneider2018} argue for the semantic value of adpositions and propose the Semantic Network of Adposition and Case Supersenses (SNACS) schema.

SNACS categorizes the use of adpositions and case markers, including English possessives, into 50 coarse-grained supersense classes. Each adposition token is annotated as a construal construction with two of these supersenses~\cite{hwang-etal-2017-double}. A construal includes a \textsc{Scene Role} and a \textsc{Function}, where the former expresses the adposition's meaning in context and the latter denotes its lexical meaning.
An example of construal is shown in \cref{ex:intro3}, a sentence from our corpus.\footnote{Examples use the notation \rf{SceneRole}{Function}. When Scene Role and Function have the same supersense label, we write it only once for conciseness.} In context, the possessive \emph{my} expresses that the speaker is a member of an organization (the company that employs them), hence a scene role of \psst{OrgMember}; however, the lexical meaning of a grammatical possessive when not indicating possession expresses a looser relationship between entities, corresponding to the function \psst{Gestalt}. Scene role and function are drawn from the same inventory of supersenses and are often identical. In the PASTRIE corpus, 72\% of annotation targets have the same scene role and function.

\ex.\label{ex:intro3}This is why \p{my}/\rf{OrgMember}{Gestalt} employer has just finished updating 50 k users \p{from}/\psst{Source} XP \p{to}/\psst{Goal} Windows 7 .

\subsection{Prepositions are uniquely challenging for learners}\label{sec:l1-influence}
Prepositions are notoriously difficult for language learners~\cite{takahaski1969,littlemore2006,mueller2012}, which is one of the motivations for constructing this corpus. In studying English preposition usage patterns of high-proficiency learners with different L1 backgrounds, we aim to learn more about how these speakers' L1s might influence their English preposition usage, and how this information might be used to improve pedagogy.
One of the problems with prepositions is that they often seem to convey less meaning than content words such as nouns or verbs, but at the same time can be nuanced and highly polysemous. \citet{erarslan2014language} observed that ``most L1 interference took place in the use of prepositions and vocabulary following it.''\footnote{In second language acquisition, the terms \emph{crosslinguistic influence}, \emph{interference}, and \emph{transfer} refer broadly to the characteristics of a speaker's use of a second language that can be attributed to entrenched patterns from their native language \citep{jarvis-13}.} \citet{Nacey_Graedler_2015} found rates of inappropriate preposition choice of 4--5\% (out of all prepositions) in two corpora of advanced English speakers with a Norwegian L1. They found learners' oral production as challenging as written production and their analysis of the International Corpus of Learner English \cite[``ICLE'';][]{granger2009international} provided evidence of speakers' L1s influencing L2 lexical choice in both positive and negative ways.

\citet{mahmoodzadeh2012} conducted Persian-English translation task-based experiments focused on identifying preposition error types. He found that the intermediate Iranian learners of English made more errors of redundancy or inappropriate use than errors of omission and discussed several transfer-related causes of these errors. \citet{gvarishvili2013} explored negative L1 interference in English preposition usage and offered advice to language educators for mitigating it, but also suggested that educators take advantage of positive influence by pointing out to students L1 prepositions with similar use as their English counterparts.

The difficulty of acquiring prepositions when learning a new language is also addressed in cognitive studies. \citet{lowie2004} offered a cognitive discussion of the progression of preposition acquisition in Dutch learners of English; \citet{hung2018} found a cognitive approach that focuses on both spatial and metaphorical meanings to be effective for teaching English prepositions; and \citet{tyler2012,bratoz2014,wong2018,zhao2020} all advocate for a cognitively driven approach to teaching prepositions as well. Pedagogical approaches for teaching English prepositions are also compared in \citet{mueller2011, mueller2012}. Given the particular difficulty of preposition acquisition, these cognitive pedagogical approaches and new insights from studying learner data should be put to use to help students. 

Automatic grammatical error detection (and correction) is another tool that can aid students and has been widely studied, including for prepositions specifically. Models of native and non-native English have been used to predict and detect preposition errors \cite{chodorow-07,de_felice-08,tetreault-08,hermet-09,tetreault-10,gamon-10,barak-20}. \Citet{graen-17} took advantage of parallel corpora for identifying challenging prepositions for learners; \citet{madnani-11} proposed a crowdsourcing-based approach for improving evaluation of grammatical error detection systems; and \citet{huang-16} built a Chinese preposition selection model to aid in identifying errors and correcting them. 
Recently, \citet{nagata_creating_2020} developed a corpus of explicit textual feedback for preposition errors in essays by Asian learners of English, while \citet{farias_wanderley-21} provided similar annotations of several classes of errors including a judgment of whether each error was due to transfer (from L1 Chinese). 
Making precise some notion of preposition \emph{meaning}, as we do with the PASTRIE corpus, holds the potential to give more systematic corrective feedback regarding meaning.
But we stress that differences between L1 and L2 preposition usage are not limited to \emph{errors} made by \emph{learners}; L1-influenced statistical tendencies in fluent usage are of interest as well.

\subsection{Reddit-L2: Our Source Corpus}
The Reddit-L2 corpus was published in 2018 alongside an analysis of cognate effects in language produced by non-native speakers of English \cite{rabinovich2018}. It contains 230M sentences and 3.5B tokens of English data from native and non-native speakers, whose L1s were heuristically identified. It was created by first selecting users with a self-specified country \emph{flair} on a set of subreddits and then gathering additional content from those users on different subreddits. While knowing a user's country does not guarantee that their L1 is the majority language in that country, steps were taken to make this more likely, and the inherent noise in the data is acknowledged in the corpus description.\footnote{Further details on the construction of the Reddit-L2 corpus are available in section 3 of \citet{rabinovich2018}.} The corpus focuses on large languages (it includes authors with flairs from 31 countries representing the Germanic, Romance, and Balto-Slavic language families) and excludes multilingual countries like Switzerland.

Since being made available, the corpus has primarily been used for native language identification \cite{goldin2018,kumar2019,steinbakken2019,sarwar2020}. However, it has also been used in studies of bias in word embeddings and bias against non-native text \cite{manzini2019,zhiltsova2019}, as well as semantic infelicity detection \cite{rabinovich2019}.

\subsection{An example where transfer might be expected}\label{sec:spanish-example}

Although many analyses are possible, both before and after examining the corpus, consider sentences (4)-(6) in Spanish and their English equivalents below. In these examples the supersense \psst{Locus} is realized as \textit{in}, \textit{on}, and \textit{at} in English, whereas in Spanish, the locative meaning is expressed with \textit{en} across contexts.

\ex. 
\begin{enumerate}
    \item[]  Al est\'{a} en Chicago.
    \item[]  Al is \textbf{in} Chicago.
\end{enumerate}

\ex.
\begin{enumerate}
    \item[] Al est\'{a} en la playa.
    \item[] Al is \textbf{on} the beach.
\end{enumerate}

\ex.
\begin{enumerate}[label=(\alph*)]
    \item[]  Al est\'{a} en la fiesta.
    \item[]  Al is \textbf{at} the party.
\end{enumerate}

The examples from Spanish motivate the following hypothesis: in a corpus of L2 English writing, L1 speakers of Spanish will overgeneralize to the English preposition \textit{in} because of perceived similarity with the L1 form \textit{en} when conveying related spatial meanings. 
Part of the L2 acquisition challenge is the conceptual grounding of spatial language, as conceptual space is organized differently from language to language and 
is acquired early by children \cite{bowerman-01,hall2018cup}.

Evidence of L1-influenced use of a single preposition can take two forms. In some cases \emph{overuse} can lead to an overt error, or wholesale exchange of a preposition as in \textit{Al was in the beach} to mean that Al had been sunning on the beach, or, that Al was at the beach. The second source of possible L1 influence is \emph{avoidance} of using a particular preposition, demonstrated by lower relative use. Taken together, it is proposed that L1 Spanish writers will demonstrate more frequent use of \textit{in} during production, in those cases where the supersense \psst{Locus} is indeed distributed across the other locatives available in the L2. Prior to analyzing the corpus, it was reasonable to predict that L1 Spanish writers would demonstrate the highest frequency use of the preposition \textit{in}, and the lowest usage of the preposition \textit{at}.

\section{Corpus Description}

\subsection{The PASTRIE Corpus}\label{sec:description}

The PASTRIE corpus consists of 1,155 sentences and 22,484 tokens from 255 Reddit documents sampled from the following languages and countries, with percentages of tokens in parentheses:

\begin{itemize}
    \item English (24.07\%): Australia, New Zealand, UK, US
    \item French (23.56\%): France
    \item German (28.08\%): Austria, Germany 
    \item Spanish (24.29\%): Argentina, Mexico, Spain
\end{itemize}

English was chosen as a baseline for comparisons, and French, German, and Spanish were chosen due to their relative similarity to English and wide availability in the Reddit-L2 corpus. While it's possible that some documents belong to the same Reddit thread, this was not a specific selection criterion. 

In the corpus, there are 2,395 annotation targets. Of these targets, 2,193 are single tokens and 202 are prepositional MWEs. Sentence segmentation and tokenization were performed with StanfordNLP \cite{qi2018}, and annotation targets, including prepositions, possessives, MWEs, and infinitives, were identified heuristically using the same script used for the STREUSLE corpus \cite{schneider2018} and then manually corrected during annotation.

\subsection{Annotation}

\subsubsection{Annotation Process}
We organized the annotation effort into smaller samples of data (annotation ``tasks'') that each included 15 documents, and we annotated a total of 17 tasks. All tasks were assigned documents randomly, and documents of each L1 appeared in each task. Tasks were independently annotated by two different annotators, then adjudicated in a meeting which included both annotators and at least one additional person who led the adjudication. Annotators and adjudicators were not shown the L1s of specific documents.

Four different annotators participated over the course of the project, all of whom were Linguistics graduate students and native English speakers; one additional person, a professor with expertise in the annotation scheme did not annotate, but participated in adjudication meetings, especially in the early stages of the project to ensure accuracy. Over all targets, the two annotators agreed on 59.2\% of Scene roles (Cohen's $\kappa = 0.58$) and 68.2\% of Function labels ($\kappa = 0.66$). This is lower than the SNACS IAA numbers found in \citet{schneider2018} (74.4\% agreement on Scene, 81.3\% on Function); however, those were on a sample from a single text, The Little Prince; our data is likely more difficult due to the wide range of topics and authors on social media, and the use of informal and sometimes non-native language.

After initial annotation and adjudication was complete, we did an additional review to ensure annotations were consistent with version 2.5 of the guidelines \citep{snacs-guidelines}, since most annotation had been done with previous versions, and to resolve difficult cases that were initially left as open or marked as uncertain.

\subsubsection{Challenging Cases}
One challenge in annotating the data is that Reddit contains discussion of a wide range of topics, often using jargon that would be understood by members of a given subreddit but was not always familiar to annotators; in these cases, annotators looked up terminology or consulted with others to ensure they understood the sentences. The range of topics also meant that many interesting semantic relationships appeared in the data that had not been seen in the STREUSLE corpus. For example, \cref{ex:videogame} discusses the details of a video game, where a decision had to be made whether to treat \emph{the game} as personified when annotating \emph{by}.

\ex.\label{ex:videogame}Only Zin and Gore can be knocked \p{out\_of}/\psst{Source} \p{their}/\psst{Gestalt} charged modes , but those are n't considered \p{as}/\rf{Characteristic}{Identity} enraged \p{by}/\rf{Experiencer}{Agent} the game .

In some cases, adpositions represented ambiguous semantic relationships and adjudicators had to decide on the most likely interpretations. For example, in \cref{ex:recipe}, we considered whether the recipe could be considered a personified \psst{Originator} of the suggestion, in which case the possessive would be annotated \rf{Originator}{Gestalt}. We decided that while this is a possible interpretation, it was more straightforward to consider the suggestion as part of the recipe, hence \rf{Whole}{Gestalt}.

\ex.\label{ex:recipe}Never follow a recipe \p{'s}/\rf{Whole}{Gestalt} suggestion \p{for}/\psst{Topic} how much garlic you should put \p{in}/\rf{Goal}{Locus} .

Finally, as with most social media, Reddit text is largely written in an informal register with little or no editing. While this rarely posed a problem for annotation, there were some cases where it was difficult to discern the intended meaning of a sentence. For example, it is unclear whether \cref{ex:hypothesis} is referring to a hypothesis that leads to taking a measurement, or the hypothesis made based on a measurement. We decided that it was most likely either \psst{Explanation} or \psst{Purpose}, and chose \psst{Explanation} because it is more general. In \cref{ex:typo}, the preposition \emph{of} doesn't make sense; we annotated it as a typo for \emph{off}, but it could conceivably be a typo for \emph{on} instead.

\ex.\label{ex:hypothesis}You can doubt the hypothesis \p{for}/\psst{Explanation} a measurement but you can not doubt the actual measurement .

\ex.\label{ex:typo}The easiest is getting a bunch of chickens , a rooster , and live \p{of}/\rf{Instrument}{Source} eggs .

\section{Analysis}
\subsection{Preposition Usage} 
The PASTRIE corpus is an annotated subcorpus of a larger initial sample we drew from the Reddit-L2 corpus. This sample of roughly 2,500 documents is a more representative source for analysis of preposition usage and can serve as supplementary data for future annotation. 

The statistics of the initial sample are described in \cref{tab:parent} and the statistics of the annotated PASTRIE corpus are described in \cref{tab:child}. We see no alarming deviations in PASTRIE compared to the initial sample. PASTRIE contains more English tokens generated by some L1s than others as a result of the random sampling involved in task generation.

\begin{table*}[ht]
\centering\small
\begin{tabular}{|c|c|c|c|c|c|c|c|}
\hline \textbf{L1} & \textbf{Documents} & \textbf{Tokens} & \textbf{Sentences} & \textbf{Prepositions} & \textbf{Prepositions/Token} & \textbf{Tokens/Doc} & \textbf{Sentences/Doc} \\
\hline
English & 658 & 48,529 & 2,544 & 5038 & 10.28\% & 73.75 & 3.87 \\
French & 677 & 52,093 & 2,689 & 5213  & 10.01\% & 76.95 & 3.97 \\
German & 767 & 69,206 & 3,681 & 7380 & 10.66\% & 90.23 & 4.80 \\
Spanish & 587 & 45,488 & 2,410 & 4588 & 10.09\% & 77.49 & 4.11 \\
\hline
\end{tabular}
\caption{\label{tab:parent} Characteristics of the initial sample of the Reddit-L2 corpus which tasks were created from.
}
\end{table*}

\begin{table*}[ht]
\centering\small
\begin{tabular}{|c|c|c|c|c|c|c|c|}
\hline \textbf{L1} & \textbf{Documents} & \textbf{Tokens} & \textbf{Sentences} & \textbf{Prepositions} & \textbf{Prepositions/Token} & \textbf{Tokens/Doc} & \textbf{Sentences/Doc} \\
\hline
English & 67 & 5,412 & 284 & 579 & 10.70\% & 80.78 & 4.24 \\
French & 74 & 5,297 & 281 & 539 & 10.18\% & 71.58 & 3.80 \\
German & 74 & 6,313 & 334 & 675 & 10.69\% & 85.31 & 4.51 \\
Spanish & 65 & 5,462 & 256 & 602 & 11.00\% & 84.03 & 3.94 \\
\hline
\end{tabular}
\caption{\label{tab:child} Characteristics of the PASTRIE corpus, the annotated subset of data.
}
\end{table*}

As shown in \cref{tab:parent}, prepositions tend to make up 10--11\% of the data. The rate of preposition use is highest for German L1 speakers, followed by English and Spanish, with French being the lowest.\footnote{The corpus only contains English data. When we mention other languages, we are referring to the L1 of the speaker.} While German does have the highest rate of preposition use in the initial sample, the range is only 0.65\%. This widens slightly to 0.82\% for the annotated portion of the data, which has a slightly different ranking.

\begin{figure}[ht]
\centering 
\includegraphics[width=\linewidth]{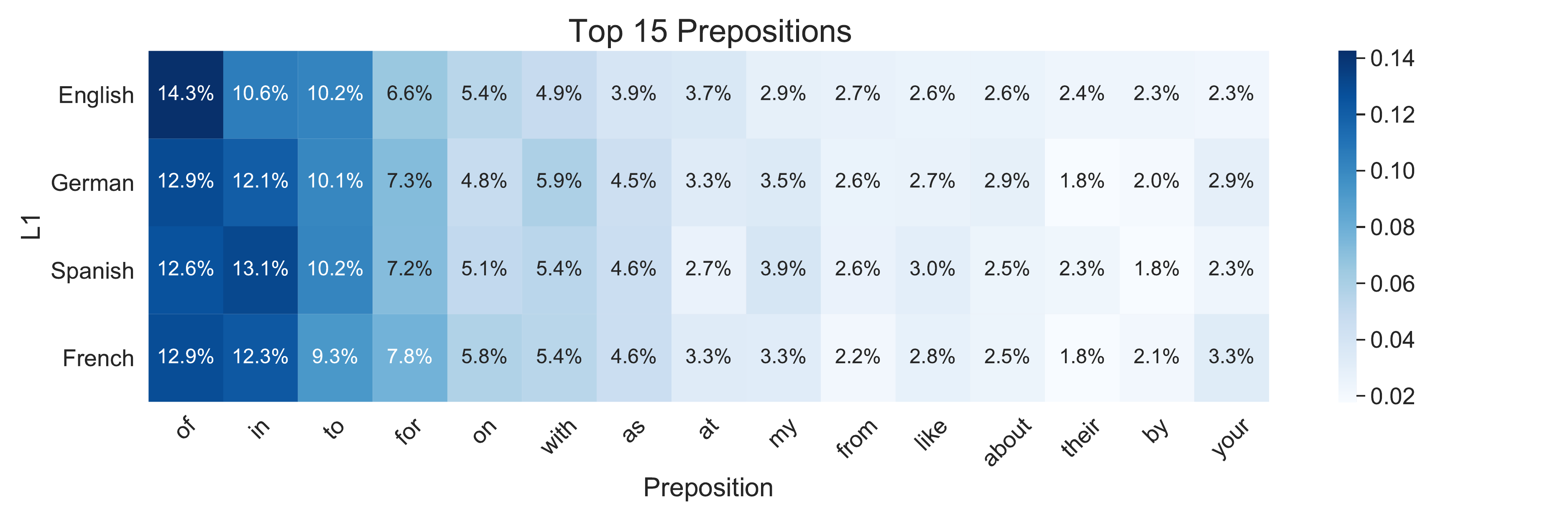}
\caption{Relative frequencies (among all prepositions for the L1) of 15 most frequent prepositions in the larger unannotated sample of the Reddit-L2 corpus.}
\label{fig:heatmap}
\end{figure}

\Cref{fig:heatmap} indicates that there are some differences in the usage of specific prepositions. Notably, \emph{of} is generated more by native English speakers than by any other speaker, whereas words like \emph{in} and \emph{with} are generated more frequently by all other L1 speakers than by native English speakers in this corpus. This may suggest that multiple senses of \emph{of} translate to distinct prepositions in other languages. This could also be due to the mechanics of possession in non-English languages.

Of the top 15 most frequently used prepositions, German L1 speakers collectively had lower relative frequency than at least one other category of L1 speaker for 13/15 prepositions, and a lower relative frequency than at least two other categories of L1 speakers for 11/15 prepositions. Broadly, this suggests that, as shown in \Cref{fig:heatmap}, German L1 speakers use the 15 most frequently used prepositions less frequently than other L1 speakers, indicating that L1 German learners of English may use a wider variety of prepositions. This could be due to prepositional transfer. The broader use of prepositions by German L1 speakers in this corpus is likely not exclusively due to increased proficiency or near-native English fluency, because German L1 preposition usage does not most closely match the preposition usage of L1 English speakers, as seen in \cref{fig:stacked_density}. These observations should be taken in context, with caveats of a small sample size and no control for topic and domain of the posts.

\subsection{Supersense Usage}

\begin{figure}[]
\centering 
\begin{subfigure}[t]{0.45\textwidth}
    \centering
    \includegraphics[width=\linewidth]{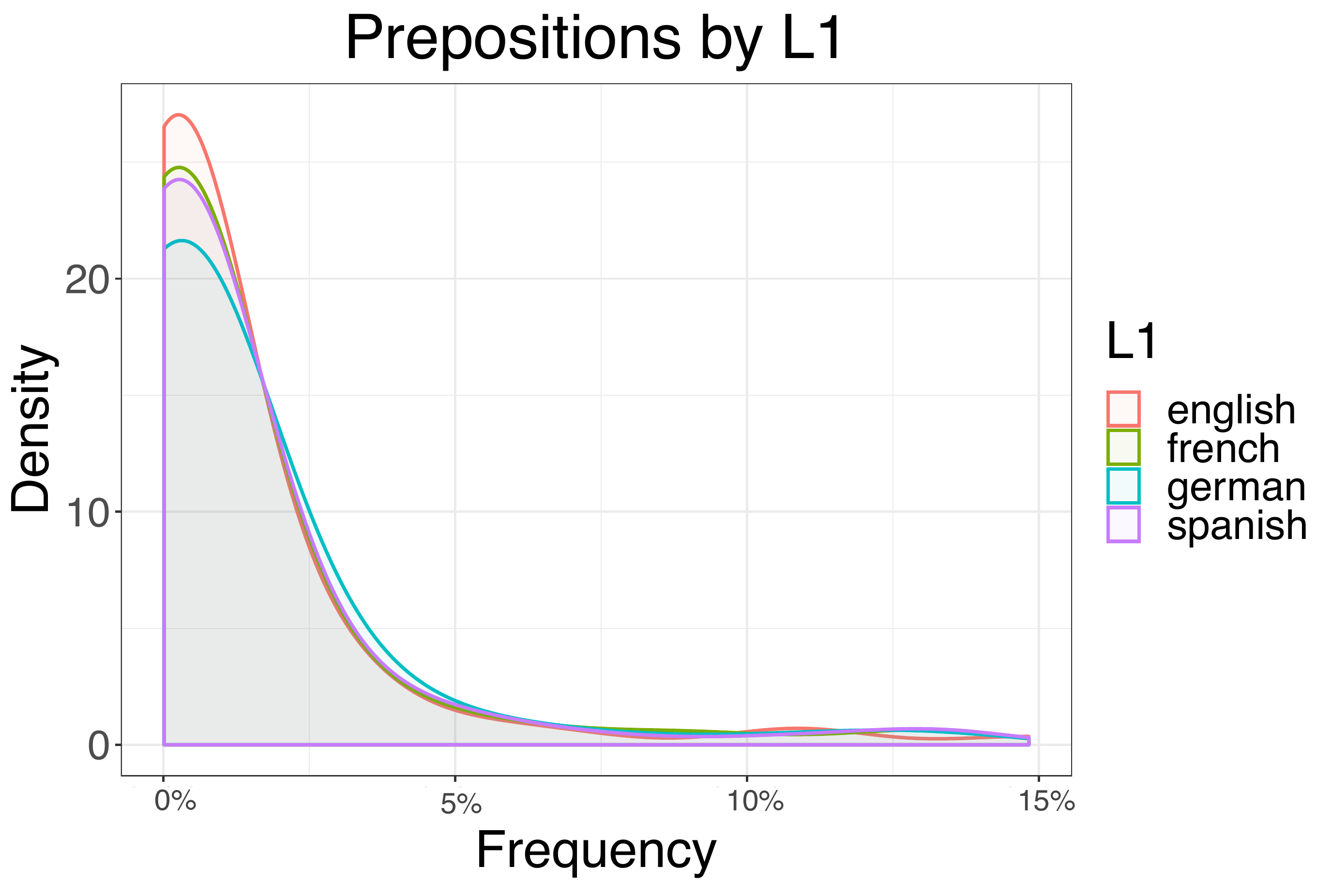}
    \caption{Density plot for preposition usage by L1, demonstrating that German has the longest tail, while English and French have the shortest tails.}
    \label{fig:stacked_density}
\end{subfigure}
\hspace{0.02\textwidth}
   \begin{subfigure}[t]{0.45\textwidth}
        \centering
       \includegraphics[width=\linewidth]{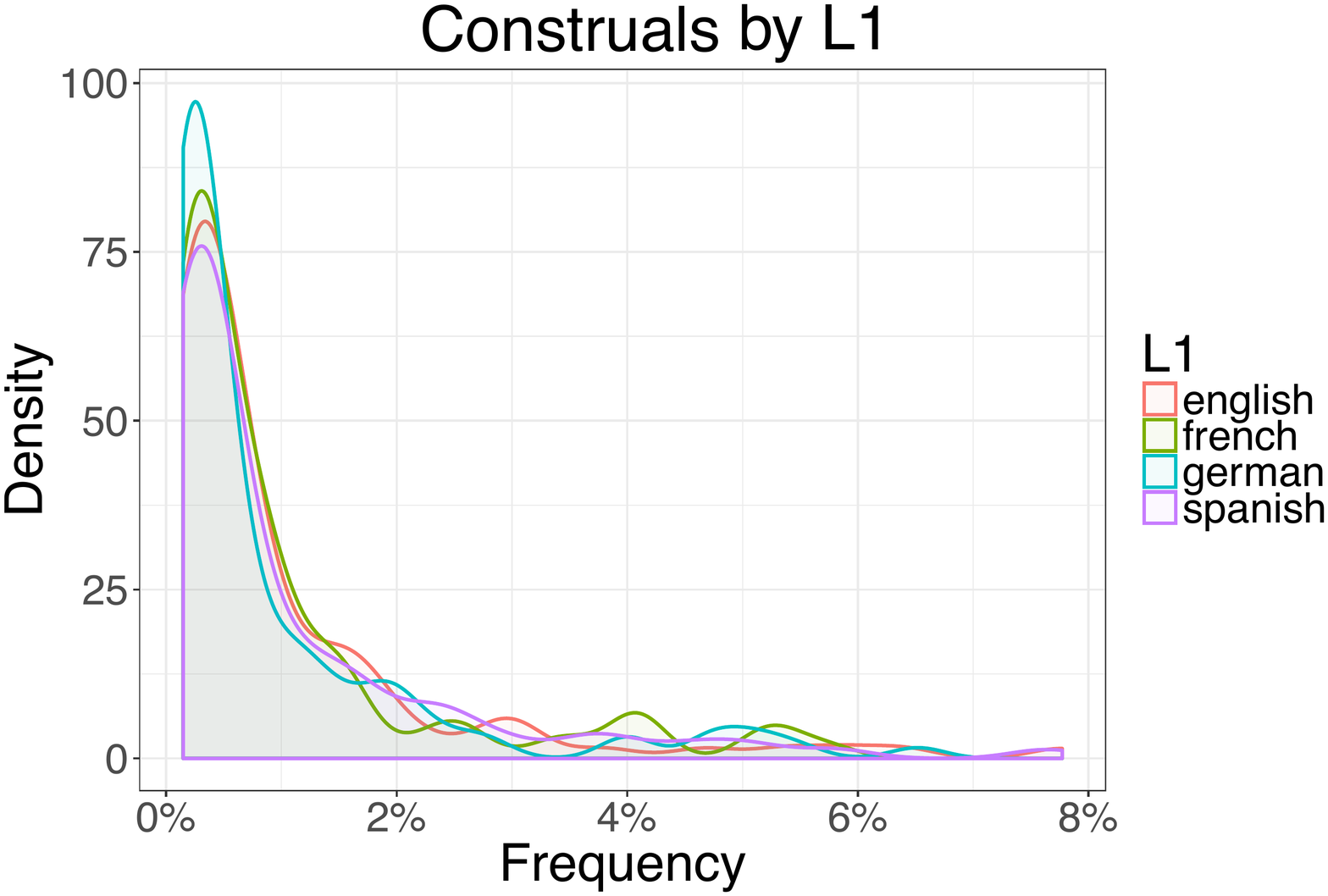}
    \caption{Density plot of construals by L1. This depicts the number of construals (y-axis) that have a certain frequency (x-axis) in the annotated corpus. Most construals are used rarely, and there are only a few high-frequency construals.}
    \label{fig:density}
    \end{subfigure}
\caption{\label{fig:density_plots} Density plots for prepositions and construals, normalized by total number of prepositions per L1. Recall that a construal is a pair of supersenses---a \textsc{Scene Role} and a \textsc{Function}.}
\end{figure}

The distributions of preposition and supersense usage by L1, depicted in \cref{fig:density_plots}, are generally comparable in shape. In the preposition usage plot, values are normalized by total frequency of prepositions for each L1. In the supersense plot, values are normalized by the total frequency of the particular construals for each L1.

\paragraph{L1 German Construal Usage} Notably, German L1 speakers have the highest number of infrequent \textit{construals}, and the lowest number of infrequent \textit{prepositions}.
German also has the longest tail when considering prepositions, while all languages have tails of similar length in the construals plot.  The peaked head close to the y-axis indicates a high number of prepositions that are used infrequently and a small number of prepositions that occur frequently. The plot also shows that a few construals and prepositions dominate usage, while most construals and prepositions are infrequently used. The density plot of German L1 construal usage has a less peaked head (smaller number of low-frequency prepositions) and a less steep decline, meaning German L1 speakers were more likely to use moderate- or high-frequency prepositions. 

This supports our claim in Section 4.1, that the range of German L1 preposition usage is being impacted by prepositional transfer. The construal usage by L1 German  speakers is not mirroring the construal usage of L1 English speakers, which would be an indication of near-native English fluency and usage, but instead presents differently than the construal usage by all three other L1 speakers.

The density values are normalized by the number of total prepositions generated by each L1, so the less peaked head is not caused by German L1 speakers having generated a larger number of prepositions.

\begin{table*}[]
    \centering \small
    \begin{tabular}{c|lr|lr|lr|lr|lr|}
    \multicolumn{3}{c}{\textbf{All}} & \multicolumn{2}{c}{\textbf{English}} & \multicolumn{2}{c}{\textbf{French}} & \multicolumn{2}{c}{\textbf{German}} & \multicolumn{2}{c}{\textbf{Spanish}} \\
    \hline
    \multirow{5}{*}{\rotatebox{90}{\textbf{Scene Role}}}
  & Locus         & 168 & \textasciigrave i            & 45 & Topic         & 35 & Topic         & 49 & Locus         & 51 \\
  & Topic         & 155 & Locus         & 41 & Theme         & 35 & Locus         & 41 & Time          & 39 \\
  & Theme         & 139 & Topic         & 39 & Locus         & 35 & CompRef. & 38 & Theme         & 38 \\
  & \textasciigrave i            & 137 & Gestalt       & 38 & Gestalt       & 30 & Gestalt       & 37 & Goal          & 36 \\
  & Gestalt       & 127 & Theme         & 37 & Circum.  & 28 & Circum.  & 35 & CompRef. & 35 \\
    \hline
    \multirow{5}{*}{\rotatebox{90}{\textbf{Function}}}
  & Gestalt       & 325 & Gestalt       & 88 & Gestalt       & 74 & Gestalt       & 87 & Gestalt       & 76 \\
  & Locus         & 242 & Locus         & 55 & Locus         & 55 & Locus         & 60 & Locus         & 72 \\
  & Topic         & 154 & Goal          & 49 & Topic         & 33 & Topic         & 51 & Topic         & 36 \\
  & Goal          & 153 & \textasciigrave i            & 45 & CompRef. & 30 & Goal          & 44 & Time          & 35 \\
  & \textasciigrave i            & 137 & Topic         & 34 & Goal          & 28 & CompRef. & 35 & Goal          & 32 \\
    \hline
    
    \multirow{5}{*}{\rotatebox{90}{\textbf{Construal}}}
  & Locus         & 146 & \textasciigrave i      & 45 & Topic        & 31 & Topic        & 44 & Locus         & 46 \\
  & Topic         & 137 & Gestalt & 37 & Locus        & 29 & Locus        & 37 & Time          & 35 \\
  & \textasciigrave i            & 137 & Locus   & 34 & Gestalt      & 28 & \textasciigrave i           & 35 & Topic         & 31 \\
  & Gestalt       & 121 & Topic   & 31 & \textasciigrave i           & 28 & Gestalt      & 34 & \textasciigrave i            & 29 \\
  & Time          & 106 & Goal    & 27 & \textasciigrave d           & 22 & Circum. & 32 & Theme         & 27 \\
    \hline
    & \textbf{Total} & 2395 & \textbf{Total} & 579 & \textbf{Total} & 539 & \textbf{Total} & 675 & \textbf{Total} & 602 \\
    \hline
    \end{tabular}
    \caption{\label{table:supersenses} Top scene roles, functions, and construals by L1. In all of the most common construals, the scene role matches the function, so only one supersense is shown.}
\end{table*}

\paragraph{Most frequent supersenses} \Cref{table:supersenses} shows that the top labels (for scene role, function, and construal as a whole) across all L1s draw from a limited set of supersenses. The top function supersense across all languages is \psst{Gestalt}, which is a prototypical function of the English genitives: \textit{of}, \textit{'s}, and the various pronominal forms. \citet{schneider2018} formulated the SNACS guidelines for these as they were very frequent in past annotated corpora and are highly polysemous; both of these attributes are evident in PASTRIE.

\psst{Locus} is the second most common function in all of the languages. Another example of variation is English's relatively high use of \olbl{\backtick i}, the infinitival uses of \textit{to} and \textit{for}, which are idiomatic to English and thus more difficult to acquire for L2 speakers \citep{heil2019acquisition}.

\paragraph{Supersense distribution comparison} \Cref{tab:jaccard} is a pairwise quantitative comparison of the construals that were encountered between each L1. It shows slight variation in scene role and function and almost no variation in the distribution of construal usage between every pair of languages. This suggests that the general set of meanings that are filled by prepositions is not substantially affected by the L1 of the speaker. Rather, we find that differences manifest in preposition choice for specific construals. An instance of variation in preposition choice for \psst{Locus} is examined below.

\begin{table}[]
    \centering
       \begin{tabular}[width=\linewidth]{l@{ \small{vs.} }l|rrr}
        \textbf{L1} & \textbf{L1} & \textbf{Scene} & \textbf{Fxn.} & \textbf{Cons.} \\
        \hline
        English & German & 0.71 & 0.73 & 0.61 \\
        English & French & 0.71 & 0.76 & 0.61 \\
        French & Spanish & 0.70 & 0.76 & 0.59 \\
        English & Spanish & 0.70 & 0.73 & 0.61 \\
        French & German & 0.69 & 0.71 & 0.60 \\
        German & Spanish & 0.67 & 0.72 & 0.61
        \end{tabular}
        \caption{\label{tab:jaccard} Jaccard similarity coefficients of the multisets of scene roles, functions, and construals between every language pair. Jaccard similarity is a metric of similarity between two sets $A$, $B$, defined as $\frac{|A \cap B|}{|A \cup B|}$.}
\end{table}

\paragraph{Variation in \psst{Locus} prepositions}

When the data is examined more narrowly, we do find examples of L1 influence on preposition choice. The most common prepositions used to represent the scene role of \psst{Locus} are \emph{in} and \emph{on}. In the British National Corpus \cite[2007]{bnc2007}, which draws from both formal and informal, written and spoken sources of English, for every instance of \emph{in} there are only 0.35 instances of \emph{on}. In the entire PASTRIE corpus, we find 0.44 instances of \textit{on} for each \textit{in} (disregarding supersense labels).

\begin{figure}[]
\centering
\centering
\includegraphics[width=0.5\linewidth]{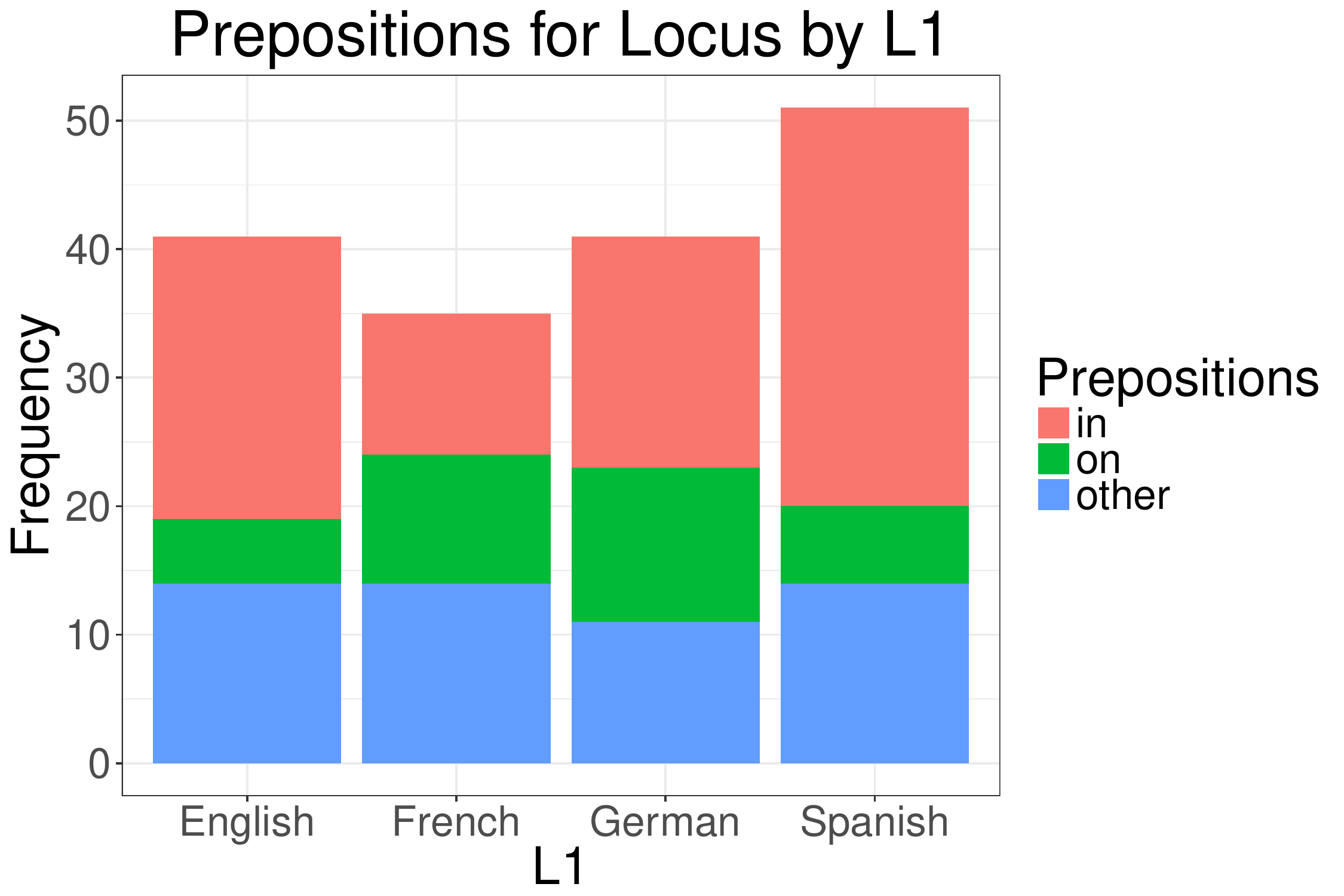}
\caption{Frequency counts of tokens annotated with \psst{Locus} as the scene role, broken down by native language and preposition type: \emph{in}, \emph{on}, and others.}
\label{fig:locus_preps}
\end{figure}

However, we find that the L1 of non-native speakers substantially skews this ratio when only considering spatial uses of these prepositions. Figure \ref{fig:locus_preps} shows that the rate of \psst{Locus} use of \emph{on} relative to \emph{in} is much greater for French (0.91) and German (0.67) than for English (0.23) and Spanish (0.19).

Previous work has observed that spatial relations are categorized differently across languages \citep{bowerman-01,feist2008space} and have complex semantics \citep{feist2000and}. In English, \emph{in} and \emph{on}, both highly polysemous prepositions, further serve a variety of spatial and metaphoric non-spatial roles \citep{rice1992polysemy} which can be difficult for non-native speakers to learn. Language acquisition in regards to motion events between satellite-framed and verb-framed languages is known to be hindered by typological differences \citep{hickmann2010typological}, and more generally due to differences in the semantic fields of spatial markers \citep{reshoft2013use}.

The most likely explanation for these discrepancies across prepositions for \psst{Locus} across L1s is that the semantic fields of spatial markers used in the L1 influences the use of those in the L2. 
L1~Spanish speakers' greater preference for  \emph{in} over \emph{on} to express locative meanings is consistent with the prediction of transfer outlined in \cref{sec:spanish-example}. 
\citet{vsevskauskiene2020prepositions}, from a pedagogical standpoint, find transfer in Lithuanian L1 speakers' acquisition of English---\textit{in} is learned more readily because it has a clear equivalent in Lithuanian's locative case, while \textit{on} is more difficult because it lacks such an equivalent. \citet{johannes2016lexicalverbs} also examine spatial prepositions in the context of L1 English acquisition, finding that in children the semantic field of \textit{on} is learned much later than that of \textit{in}.

It is possible that the cross-L1 variation in this dataset is due at least in part to factors apart from L1, such as varying topics and domains (i.e. which subreddits the documents were sampled from). Nevertheless, this example illustrates the utility of SNACS for exploratory analysis in examining and comparing adposition and case semantics.

\section{Conclusion}
With data drawn from the Reddit-L2 corpus \cite{rabinovich2018}, we created PASTRIE, a new corpus of preposition supersense annotations that is publicly available. This corpus adds to existing resources with preposition supersenses and includes annotations of native English data and data produced by L1 speakers of French, German, and Spanish. We demonstrated the applicability of SNACS and the construal analysis to L2 English. We presented detailed discussion of the annotation process, general corpus statistics, and an analysis of usage phenomena across the L1s, including variations between speakers of different L1 backgrounds.

Future work may consider a wider variety of L1s than the typologically similar and closely genetically related languages examined in this work. Computational applications of PASTRIE in natural language understanding (NLU) of non-native English merit further investigation. Finally, corpus-based research such as in this paper can be used to empirically investigate theories of language acquisition.

\section*{Acknowledgments}
We thank Shuly Wintner, Ella Rabinovich, and Liat Nativ for their assistance in sampling data from the Reddit-L2 corpus, and we thank members of the NERT lab for their ideas and suggestions. We are grateful to Tripp Maloney, Ryan Mannion, and Sasha Slone for their careful annotation work, and to anonymous reviewers for their feedback. This research was supported in part by NSF award IIS-1812778 and grant 2016375 from the United States--Israel Binational Science Foundation (BSF), Jerusalem, Israel.

\bibliographystyle{acl_natbib}
{\smaller[.4]\bibliography{pastrie}}

\end{document}